\documentclass{article}
\usepackage{spconf,amsmath,graphicx}

\usepackage{hyperref}
\usepackage[normalem]{ulem}
\useunder{\uline}{\ul}{}
\usepackage{amsfonts}
\usepackage[sort]{cite}
\usepackage{balance}
\usepackage{wrapfig,lipsum,booktabs}

\usepackage{multirow}

\title{Exploring Latent Cross-Channel Embedding for Accurate 3D Human Pose Reconstruction in a Diffusion Framework}
\name{Junkun Jiang,~Jie Chen\thanks{The research was partially supported by the Theme-based Research Scheme, Research Grants Council of Hong Kong (T45-205/21-N), and by the HKBU Tier 2 Start-up Grant (RC-OFSGT2/20-21/SCI/001)}}
\address{Department of Computer Science,\\
Hong Kong Baptist University, Hong Kong SAR}
\begin{document}
%
\maketitle
\begin{abstract}
Monocular 3D human pose estimation poses significant challenges due to the inherent depth ambiguities that arise during the reprojection process from 2D to 3D. Conventional approaches that rely on estimating an over-fit projection matrix struggle to effectively address these challenges and often result in noisy outputs. Recent advancements in diffusion models have shown promise in incorporating structural priors to address reprojection ambiguities. However, there is still ample room for improvement as these methods often overlook the exploration of correlation between the 2D and 3D joint-level features.
In this study, we propose a novel cross-channel embedding framework that aims to fully explore the correlation between joint-level features of 3D coordinates and their 2D projections. In addition, we introduce a context guidance mechanism to facilitate the propagation of joint graph attention across latent channels during the iterative diffusion process. To evaluate the effectiveness of our proposed method, we conduct experiments on two benchmark datasets, namely Human3.6M and MPI-INF-3DHP. Our results demonstrate a significant improvement in terms of reconstruction accuracy compared to state-of-the-art methods. The code for our method will be made available online for further reference.

\end{abstract}

\section{Introduction}\label{sec:introduction}

The 3D Human Pose Estimation (HPE) technique is essential, as it encompasses wide-ranging applications in security surveillance, film making as well as emerging trends such as dance generation~\cite{au2022choreograph} and pedestrian behaviour prediction~\cite{papaioannidis2023fast}. Its primary objective is to predict the 3D locations of human body parts based on 2D observations and then reconstruct human motion. 
Despite the considerable progress on multi-view 3D HPE~\cite{jiang2022dual,zhou2022quickpose,pavlakos2019expressive,gong2022posetriplet}, monocular pose estimation remains challenging because of the depth ambiguity caused by uncertain 2D-to-3D projection and occlusion of body parts. 

To deal with the depth ambiguity, researchers mainly rely on a two-stage pipeline, i.e., utilize a 2D pose detector to obtain 2D poses, and then regress a 2D-to-3D lifting network which learns one specific view's projection implicitly~\cite{xu2021graph,zhao2022graformer,pavllo20193d,zhang2022mixste,du2023flowpose}. 
Specifically, Xu et al.~\cite{xu2021graph} propose to utilize Graph Convolutional Network (GCN) as both the encoder and decoder within the stacked hourglass network architecture for the purpose of exploring the humanoid kinematic structure. 
Following \cite{xu2021graph}, Zhao et al.~\cite{zhao2022graformer} propose to involve the attention mechanism into the vanilla GCN blocks. Therefore the attention-involved GCN block is able to model more implicit relations. 
On the other hand, many works focus on incorporating 2D motion sequences to extract spatial-temporal information from them. Pavllo et al~\cite{pavllo20193d} predict the centre frame's 3D pose from a 2D motion sequence by a dilate Convolutional Neural Network (CNN). Zhang et al.~\cite{zhang2022mixste} propose a Transformer-based model for sequence-to-sequence lifting which efficiently models skeletal structure and kinematic correlation. Du et al.~\cite{du2023flowpose} introduce Normalizing Flow to mine the feasible motion distribution across spatial and temporal domains. Despite multi-frame estimation frameworks~\cite{pavllo20193d,zhang2022mixste,du2023flowpose} leveraging temporal information to effectively address ambiguity, those frameworks rely on future frames that hinder real-time performance. In contrast, this paper focuses on a more challenging task, i.e., monocular single-frame 3D HPE.


\begin{figure*}[htbp]
\centering
\includegraphics[width=0.9\linewidth]{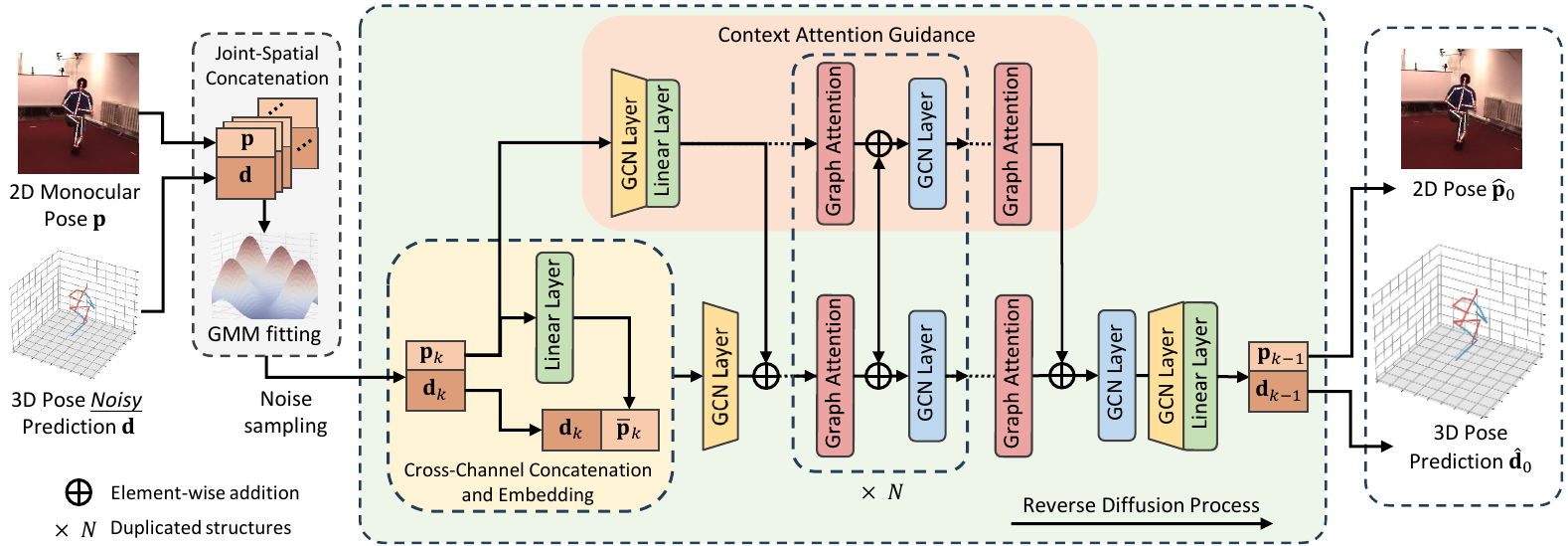}
\vspace{-10pt}
\caption{System diagram for the proposed framework during inference. 
The distribution of the 2D and  initial 3D pose predictions are fitted using a Gaussian Mixture Model, based on which $h_K= \{\mathbf{p}_K, \mathbf{d}_K\}$ will be sampled and go through $K$ iterations of reverse diffusion process until the high quality 3D pose $\hat{\mathbf{d}}_0$ is predicted. }
\label{fig:system}
\vspace{-13pt}
\end{figure*}

Recently, Denoising Diffusion Probabilistic Models~\cite{song2021denoising,ho2020denoising,gong2023diffpose,choi2022diffupose} (DDPMs, a.k.a., diffusion models) have garnered the interest of researchers owing to their high-quality noise-to-image generation. Some \cite{gong2023diffpose,choi2022diffupose} propose to regard the depth ambiguity as the noise and utilize the diffusion model to enhance 2D-to-3D lifting results. In particular, Choi et al.~\cite{choi2022diffupose} suggest replacing the standard DDPM's U-Net structure with a GCN. They input noisy 3D joints, initialized with the Gaussian distribution and condition the 3D noise on 2D observation. Similarly, Gong et al.~\cite{gong2023diffpose} also adopt a GCN structure~\cite{zhao2022graformer} and accelerate the diffusion process using DDIM~\cite{song2021denoising}. They assume the noisy 3D input is sampled from an indeterminate pose distribution.
Inspired by ~\cite{gong2023diffpose, choi2022diffupose}, we propose a monocular single-frame 3D HPE framework based on diffusion modelling. Unlike the previous works, we model the correlation between both 2D and 3D uncertainty, while also incorporating the 2D-to-3D context attention. 
To be more specific, the contributions of this paper are three-fold: 1) We develop a novel monocular single-frame 3D HPE framework driven by the diffusion model that works as a post-processing to refine the lifting results. 2) Inside the framework, we first propose the Cross-Channel Embedding (CCE) module that explores the correlation between both 2D and 3D distributions. We further propose the Context Attention Guidance (CAG) module to encourage efficient cross-joint attention propagation. 
3) The proposed framework achieves state-of-the-art performance on two widely-used benchmarks~\cite{ionescu2013human3,mehta2017monocular} demonstrating our superiority.

\vspace{-5pt}
\section{Proposed Method}

Given a 2D human pose observation $\mathbf{p} \in \mathbb{R}^{2 \times J}$ and its corresponding lifted 3D prediction $\mathbf{d} \in \mathbb{R}^{3 \times J}$, our goal is to fully utilize $\mathbf{p}$ as observation context guidance and explore its correlation with the noisy low-quality prediction $\mathbf{d}$ to obtain a high-quality determinate sample $\mathbf{d'} \in \mathbb{R}^{3 \times J}$ via a diffusion probabilistic denoising process. Here $J$ denotes the total number of joints. In this section, we first give the preliminary of the diffusion modelling for HPE in Sec.~\ref{sec:background} and then describe the details of our proposed model in Sec.~\ref{sec:method}.

\vspace{-5pt}
\subsection{Preliminary for Diffusion Modelling}\label{sec:background}

Introduced in \cite{ho2020denoising,song2021denoising}, diffusion models are probabilistic generative models via modelling the transition probabilities between the Gaussian noise $h_{\mathit{K}}\sim \mathcal{N}(\mathbf{0},\mathbf{I})$ to a determinate sample $h_{0}$ via iteratively denoising from $h_{\mathit{K}}$, here $K$ denotes the total diffusion steps.
While the denoising process can be regarded as \textit{reverse diffusion}, the noise adding process, a.k.a. \textit{forward diffusion} can be formulated as:
\begin{equation}
q(h_k\mid h_0) = \mathcal{N}(h_k; \sqrt{\alpha_k}h_0,(1-\alpha_k)\mathbf{I}),
\end{equation}
which indicates that noised samples $\left \{ h_k \right \}_{k=1}^{K-1}$ can be calculated with a re-parameterized form:
\begin{equation}
\label{eq.alpha}
h_k=\sqrt{\alpha_k}h_0+\sqrt{1-\alpha_k}\epsilon,
\end{equation}
where $\epsilon $ indicates Gaussian white noise. According to Eq.~(\ref{eq.alpha}), we can directly blend the source sample $h_0$ with white noise to calculate $h_K$, and $h_K$ will become Gaussian noise eventually when $\alpha_{K} \simeq 0$. More detailed mathematical derivations can be found in \cite{song2021denoising,ho2020denoising}. 
The \textit{reverse diffusion} is usually implemented using a neural network $g_\theta$ ($\theta$ indicates learnable parameters) optimized by the source and noised sample pairs generated by the \textit{forward diffusion} process. Specifically, \cite{gong2023diffpose,choi2022diffupose} assume $h_k$ is conditioned by $\mathbf{p}$ and $\mathbf{d}$.
\cite{gong2023diffpose} assumes $h_K$ is not sampled from a Gaussian white noise.

\subsection{The proposed model}\label{sec:method}

As shown in Fig.~\ref{fig:system}, inspired by the design of the lightweight GrapFormer~\cite{zhao2022graformer}, our model takes an initial 2D and 3D pose prediction as input. 
A Cross-Channel Embedding (CCE) module is designed to facilitate a comprehensive exploration of the correlation between joint-level features of 3D coordinates and their 2D projections. These features subsequently go through $N$ stacked Attention-GCN layers ($N$ Empirically set to 5), in parallel with a Context Attention Guidance (CAG) module which further encourages efficient cross-joint attention propagation among the latent channels throughout the iterative diffusion process, until the high-quality 3D pose is estimated by the final GCN prediction layer. 

\subsubsection{Cross-Channel Embedding Module}
\begin{figure}[t]
\centering
\includegraphics[width=0.98\linewidth]{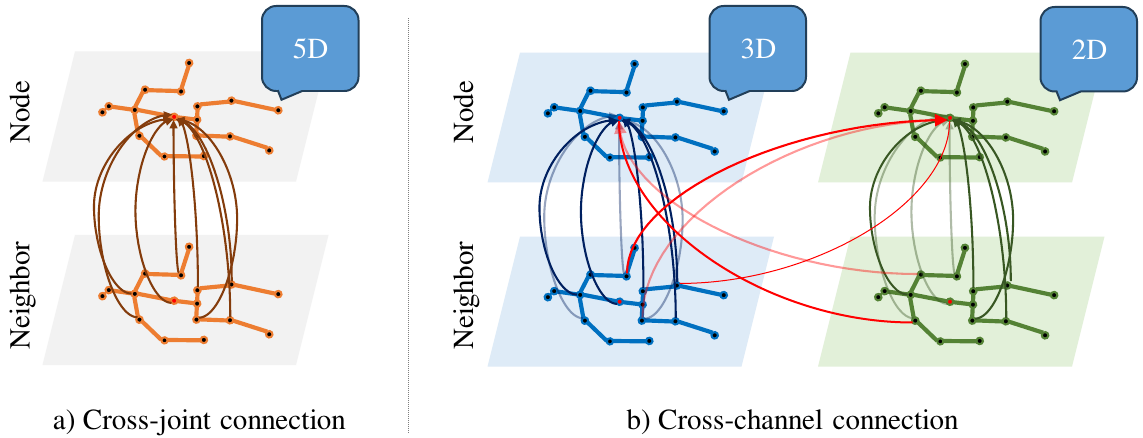}
\vspace{-10pt}
\caption{Illustration of information transfer path between node (\textit{mid-hip}) and its neighbour in a GCN. Each curve demonstrates the pathway. Left: Cross-joint connection mentioned in \cite{gong2023diffpose,choi2022diffupose}. They only consider the cross-joint correlation. Right: Cross-channel connection generated by the proposed CCE module. Ours investigates the relations between 3D coordinates and 2D projections (highlighted by red curves).}
\label{fig:channel-wise}
\end{figure}

In order to alleviate the depth uncertainties of the lifted 3D coordinates, we propose the Cross-Channel Embedding module to fully explore the correlation between the joint-level features with their 2D projections as shown in Fig.~\ref{fig:channel-wise}. Recent works~\cite{gong2023diffpose,choi2022diffupose} directly concatenate the 2D and 3D joint coordinates as $h_k$, which can be formulated as:
\begin{equation}
h_k=[ \mathbf{p}_k^T, \mathbf{d}_k^T]^T\in \mathrm{R}^{5 \times J},
\end{equation}
where $T$ is the transpose operator, and $[\cdot,\cdot]$ denotes column-wise matrix concatenation. Next $h_k$ is then embedded via GCN layers through a joint-wise adjacency matrix $\mathbf{A} \in \mathrm{R}^{J \times J}$. 
As shown in Fig.~\ref{fig:channel-wise} (a), it is apparent that only \textit{cross-joint} correlations are explored in such a framework; leaving the implicit relations between the 2D and lifted 3D features both within and across different joints under-investigated. 

Therefore, instead of direct spatial concatenation, we concatenate the 2D and 3D features along their channel dimensions, which can be formulated as:
\begin{equation}
h_k=\left[ L_e(\mathbf{p}_k), \mathbf{d}_k \right]\in \mathrm{R}^{3 \times 2J}.
\end{equation}
Where $L_e:2\times J\rightarrow3\times J$ denotes a linear projection layer that aligns the 2D and 3D features to the same dimension so that $\bar{\mathbf{p}}_k= L_e(\mathbf{p}_k) \in\mathrm(R)^{3\times J}$.
Consequently, the GCN adjacency matrix, with the extended shape of $\mathbf{A'} \in \mathrm{R}^{2J \times 2J}$, would evaluate on the connections beyond the joint-level granularity, but per 2D and 3D lifted predictions granularity instead. 
Fig.~\ref{fig:channel-wise} (b) illustrates such an example of interconnected relationships among 2D and 3D joint features. 

\subsubsection{Context Attention Guidance Module} 
As shown in Fig.~\ref{fig:system}, the proposed diffusion model consists of two branches, i.e., the signal diffusion branch, which refines the joint coordinates from noisy to high-quality; and the Context Attention Guidance branch,  which propagates cross-joint attention among the latent channels. Similar to the signal diffusion branch, the 2D coordinate features are first projected to the latent space by a GCN layer in the CAG branch, and then a linear layer aligns the feature dimension to be the same as the diffusion branch's GCN layer output. The features of the two branches are merged after the self-attention layer for Attention communication. 

\subsubsection{Training Pipeline over Forward diffusion} 
Although $h_K$ is a noisy human pose prediction with potential errors induced by depth ambiguities, it is certainly different from white Gaussian noise, of which normal diffusion models will consider as the final outcome of the diffusion process. Therefore, similar to \cite{gong2023diffpose}, we adopt a Gaussian Mixture Model (GMM) to model the distribution of $h_K$. We fit the GMM on the training set using the Expectation-Maximization (EM) algorithm to approximate the training set's 2D and 3D distributions. Thus, for every \textit{forward diffusion} step, we can use the ground truth $h_0$ to directly generate the corresponding $h_k$ sampled from the fitted GMM. Mean Squared Error (MSE) loss is used to optimize the diffusion model $g$'s parameters $\theta$, formulated as:
\begin{equation}
\mathcal{L}=\sum_{k=1}^{K}\left \| g_{\theta}(h_k, \textbf{p}_k, \textbf{d}_k) - h_{k-1} \right \|_2^2,
\end{equation}


\setlength{\tabcolsep}{4pt}
\begin{table*}
\begin{center}
\caption{Quantitative comparisons of MPJPE in millimetres (mm) on Human3.6M. The top table shows the results on ground truth 2D poses. The bottom table shows the results on detected 2D poses by CPN~\cite{li2020cascaded} detector. Bold indicates best-to-date, and underline indicates second-best for each table respectively.}
\label{table:h36m_single_gt}
\small
\begin{tabular}{lcccccccccccccccc}
\hline\noalign{\smallskip}
MPJPE (GT) & Dir. & Disc. & Eat & Greet & Phone & Photo & Pose & Pur. & Sit & SitD. & Smoke & Wait & WalkD. & Walk & WalkT. & Avg. \\ 
\hline\noalign{\smallskip}
Xu et al.~\cite{xu2021graph} & 35.8 & 38.1 & 31.0 & 35.3 & 35.8 & 43.2 & 37.3 & 31.7 & 38.4 & 45.5 & 35.4 & 36.7 & 36.8 & 27.9 & 30.7 & 35.8 \\
Zhao et al.~\cite{zhao2022graformer} & 32.0 & 38.0 & 30.4 & 34.4 & 34.7 & 43.3 & 35.2 & 31.4 & 38.0 & 46.2 & 34.2 & 35.7 & 36.1 & 27.4 & 30.6 & 35.2 \\
Gong et al.~\cite{gong2023diffpose} & 28.8 & {\ul 32.7} & 27.8 & 30.9 & 32.8 & 38.9 & {\ul 32.2} & {\ul 28.3} & {\ul 33.3} & 41.0 & 31.0 & {\ul 32.1} & 31.5 & 25.9 & 27.5 & 31.6 \\
\hline
\noalign{\smallskip}
Ours w/o guide & {\ul 28.6} & 32.8 & {\ul 26.9} & {\ul 30.7} & {\ul 32.7} & 38.5 & 32.5 & 28.4 & 33.7 & 41.1 & 31.2 & 32.2 & 31.6 & {\ul 25.4} & 26.9 & 31.5 \\
Ours w/o aug & 29.9 & 33.0 & 27.6 & 30.9 & {\ul 32.7} & {\ul 37.8} & {\ul 32.2} & 28.4 & \textbf{32.1} & \textbf{39.5} & {\ul 30.6} & {\ul 32.1} & {\ul 31.2} & 25.5 & {\ul 26.7} & {\ul 31.3} \\
Ours & \textbf{27.2} & \textbf{31.9} & \textbf{26.4} & \textbf{29.4} & \textbf{31.3} & \textbf{37.1} & \textbf{30.7} & \textbf{27.2} & \textbf{32.1} & {\ul 39.7} & \textbf{29.4} & \textbf{31.1} & \textbf{30.0} & \textbf{24.3} & \textbf{25.3} & \textbf{30.2} \\
\hline\noalign{\smallskip}

MPJPE (CPN) & Dir. & Disc. & Eat & Greet & Phone & Photo & Pose & Pur. & Sit & SitD. & Smoke & Wait & WalkD. & Walk & WalkT. & Avg. \\ 
\hline\noalign{\smallskip}
Xu et al.~\cite{xu2021graph} & 45.2 & 49.9 & 47.5 & 50.9 & 54.9 & 66.1 & 48.5 & 46.3 & 59.7 & 71.5 & 51.4 & 48.6 & 53.9 & 39.9 & 44.1 & 51.9 \\
Zhao et al.~\cite{zhao2022graformer} & 45.2 & 50.8 & 48.0 & 50.0 & 54.9 & 65.0 & 48.2 & 47.1 & 60.2 & 70.0 & 51.6 & 48.7 & 54.1 & 39.7 & 43.1 & 51.8 \\
Gong et al.~\cite{gong2023diffpose} & \textbf{42.8} & \textbf{49.1} & {\ul 45.2} & \textbf{48.7} & {\ul 52.1} & {\ul 63.5} & {\ul 46.3} & {\ul 45.2} & {\ul 58.6} & {\ul 66.3} & {\ul 50.4} & {\ul 47.6} & {\ul 52.0} & \textbf{37.6} & {\ul 40.2} & {\ul 49.7} \\
\hline\noalign{\smallskip}
Ours & {\ul 43.3} & {\ul 49.7} & \textbf{44.2} & {\ul 49.1} & \textbf{51.3} & \textbf{62.6} & \textbf{46.0} & \textbf{44.9} & \textbf{58.2} & \textbf{65.7} & \textbf{49.9} & \textbf{47.3} & \textbf{51.3} & {\ul 37.7} & \textbf{40.0} & \textbf{49.4} \\
\hline\noalign{\smallskip}
\end{tabular}
\end{center}
\vspace{-25pt}
\end{table*}

\begin{figure}[t]
\centering
\includegraphics[width=1.0\linewidth]{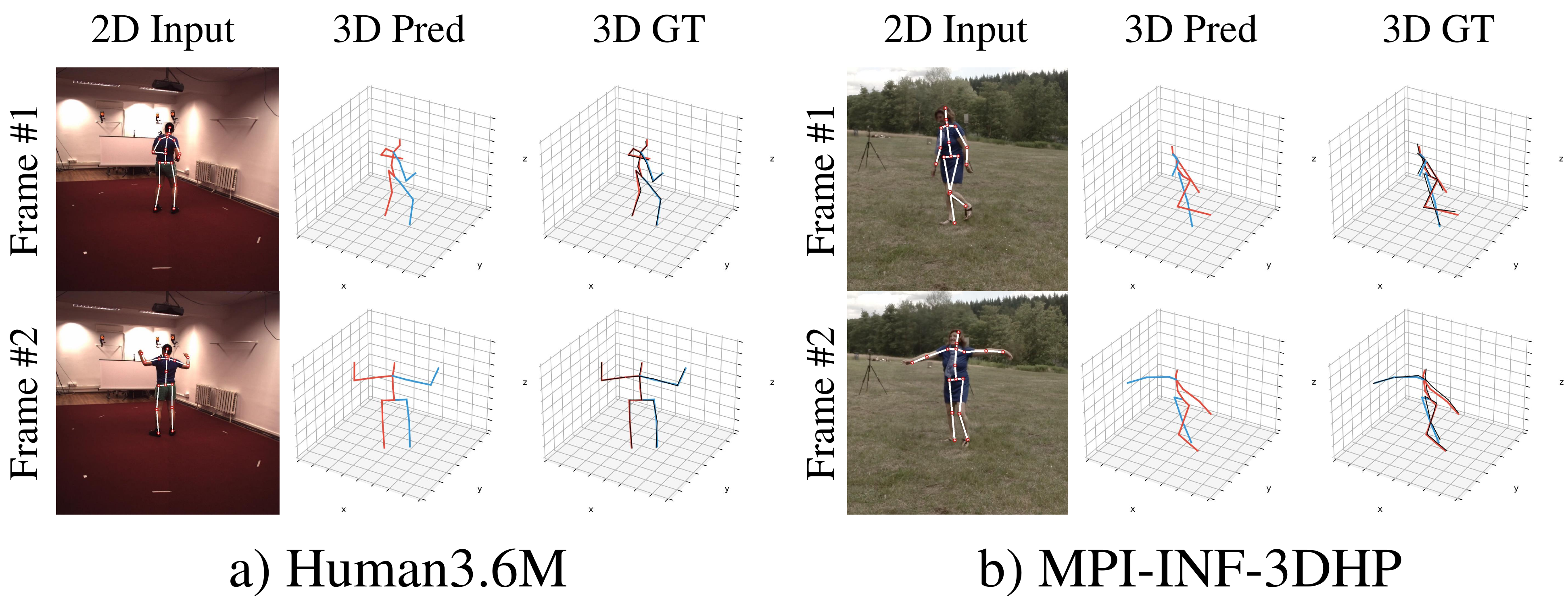}
\vspace{-20pt}
\caption{Qualitative evaluations of the proposed model trained on 2D ground truth. We provide two visual examples including the input 2D ground truth projection, 3D prediction and 3D ground truth (black) overlapped by 3D prediction.}
\label{fig:quality}
\vspace{-10pt}
\end{figure}

\setlength{\tabcolsep}{4pt}
\begin{table}
\begin{center}
\caption{Quantitative comparisons on MPI-INF-3DHP. All models are trained by 2D ground truth. $T$ denotes the number of frames. Bold indicates best-to-date, and underline indicates second-best for each table respectively.}
\label{table:mpi}
\begin{tabular}{lcccc}
\hline\noalign{\smallskip}
Method & $T$ & PCK $\uparrow$ & AUC $\uparrow$ & MPJPE $\downarrow$ \\
\hline\noalign{\smallskip}
Pavllo et al.~\cite{pavllo20193d} & 81 & 86.0 & 51.9 & 84.0 \\
Li et al.~\cite{li2022mhformer} & 9 & 93.8 & 63.3 & 58.0 \\
Zhang et al.~\cite{zhang2022mixste} & 1 & 94.2 & 63.8 & 57.9 \\
Zhang et al.~\cite{zhang2022mixste} & 27 & 94.4 & 66.5 & 54.9 \\
Gong et al.~\cite{gong2023diffpose} & 81 & \textbf{98.0} & {\ul 75.9} & {\ul 29.1} \\ 
\hline
Ours & 1 & \textbf{98.0} & \textbf{76.2} & \textbf{29.0} \\
\hline\noalign{\smallskip}
\end{tabular}
\end{center}
\vspace{-25pt}
\end{table}

\section{Experiments}

\subsection{Implementation Details}

The model is trained and evaluated on Human3.6M~\cite{ionescu2013human3} and MPI-INF-3DHP~\cite{mehta2017monocular} respectively. Human3.6M~\cite{ionescu2013human3} is a widely used dataset comprising 3.6 million images capturing 15 daily activities performed by 11 subjects. MPI-INF-3DHP~\cite{mehta2017monocular} is an outdoor dataset consisting of 1.3 million images capturing 6 outdoor activities performed by 4 subjects.
Following previous works~\cite{xu2021graph,pavllo20193d,gong2023diffpose}, for Human3.6M~\cite{ionescu2013human3}, we trained the model on five subjects (S1, S5, S6, S7 and S8) and evaluated the model on two subjects (S9 and S11). For MPI-INF-3DHP~\cite{mehta2017monocular}, we followed \cite{mehta2017monocular}'s training and testing data arrangement. 
We adopt CPN~\cite{chen2018cascaded} as the 2D pose detector. Both the 2D pose ground truth and the 2D pose prediction are used as input during inference, for comparison. Besides, the 2D poses are normalized to each camera center and the 3D poses are normalized to the mid-hip. We utilize the flipping augmentation followed by \cite{pavllo20193d}. 

The model is trained for 15 epochs with a batch size of 1024 on Human3.6M~\cite{ionescu2013human3} and 10 epochs with a batch size of 512 on MPI-INF-3DHP~\cite{mehta2017monocular} respectively. 
Following the common evaluation protocol~\cite{pavllo20193d,xu2021graph}, we report the Mean Per Joint Position Error (MPJPE), Percentage of Correct Keypoints (PCK) with the threshold of 150 mm and the Area Under the Curve (AUC) as the metrics. The GMM model is provided by \cite{gong2023diffpose}. We followed the same configuration for diffusion model training as well as the DDIM acceleration~\cite{song2021denoising} setup. 
That is, we set the number of \textit{reverse diffusion} steps $K=50$. 
The number of GMM components is set to 5.

\vspace{-15pt}
\subsection{Evaluation and Comparison with SOTA Methods}

We report the quantitative comparisons on Human3.6M~\cite{ionescu2013human3} against the state-of-the-art single-frame methods~\cite{gong2023diffpose,zhao2022graformer,xu2021graph}. Table~\ref{table:h36m_single_gt} shows MPJPE results using 2D ground truth and CPN-predicted 2D detection respectively. 
For MPI-INF-3DHP~\cite{mehta2017monocular}, Table~\ref{table:mpi} shows PCK, AUC and MPJPE results as well as the number of frames each model used. All models are trained on 2D ground-truth poses.
We highlight the best in bold and the second-best with underlines.
Meanwhile, we illustrate the qualitative results in Fig.~\ref{fig:quality}. Both the 2D ground-truth pose, 3D ground-truth pose and the diffused results are visualized.
From both tables, we can see that the proposed model outperforms all methods. Moreover,  Table~\ref{table:h36m_single_gt} shows that our model is able to regress more stable 3D reconstruction when ground truth 2D poses are provided.

\subsection{Ablation Study}
\vspace{-5pt}

We validate each proposed component by ablation experiments on the Human3.6M dataset. We utilize the 2D ground truth as input. As shown in Table \ref{table:h36m_single_gt}, the fourth, fifth and sixth columns demonstrate the ablated impact of Context Attention Guidance (denoted as \textit{Ours w/o guide}) and flipping augmentation (denoted as \textit{Ours w/o aug}) respectively. 
\begin{wraptable}{r}{4.3cm}
\vspace{-5pt}
\begin{center}
\renewcommand{\arraystretch}{1.2}
\vspace{-25pt}
\caption{Analysis of speed accelerated by DDIM~\cite{song2021denoising}.}
\label{table:speed}
\vspace{-5pt}
\small
\begin{tabular}{ccccc}
\hline
\multirow{2}{0.65cm}{DDIM Step} & \multicolumn{2}{c}{MPJPE$\downarrow$} & \multicolumn{2}{c}{FPS$\uparrow$} \\
& Ours & \cite{gong2023diffpose} & Ours & \cite{gong2023diffpose} \\
\hline
1 & 31.9 & 33.1 & 47.5 & 48.3 \\
2 & 30.2 & 31.3 & 25.7 & 26.4 \\
3 & 30.0 & 30.9 & 15.6 & 15.8 \\
4 & 29.9 & 30.7 & 11.9 & 12.0 \\ \hline
\end{tabular}
\end{center}
\vspace{-21pt}
\end{wraptable} 
Note that the effect of the Cross-Channel Embedding module can be directly referred to Gong et al.~\cite{gong2023diffpose}.
We can clearly observe the degradation of the average MPJPE in these alternative models. 

\noindent \textbf{Runtime.} We report the inference speed comparison \footnote{All evaluations are taken on an Nvidia A100 GPU, with one AMD EPYC 7302 16-Core Processor. The inference batch size is set to 1. The number of the diffusion step is set to 12. } in Table~\ref{table:speed}. We observed that the more diffusion steps, the better denoised quality. Compared to \cite{gong2023diffpose}, although there is a minor decrease in FPS, the overall prediction quality is higher. Aiming for real-time execution, the number of the DDIM step is set to 2. 

\vspace{-10pt}
\section{Conclusion}
\vspace{-5pt}

In this study, we have proposed a monocular 3D pose prediction framework. A novel cross-channel embedding module is designed which is able to fully explore the correlation between joint-level features of 3D coordinates and their 2D projections. A context guidance mechanism is proposed to facilitate the propagation of joint graph attention across latent channels during the iterative diffusion process. 
Comprehensive evaluations demonstrated a significant improvement in terms of reconstruction accuracy compared to state-of-the-art methods. 

\vfill\pagebreak
\bibliographystyle{IEEEbib}
\bibliography{myRefs}

\begin{thebibliography}{10}

\bibitem{au2022choreograph}
Ho~Yin Au, Jie Chen, Junkun Jiang, and Yike Guo,
\newblock ``Choreograph: Music-conditioned automatic dance choreography over a
  style and tempo consistent dynamic graph,''
\newblock in {\em Proceedings of the 30th ACM International Conference on
  Multimedia}, 2022, pp. 3917--3925.

\bibitem{papaioannidis2023fast}
Christos Papaioannidis, Ioannis Mademlis, and Ioannis Pitas,
\newblock ``Fast single-person 2d human pose estimation using multi-task
  convolutional neural networks,''
\newblock in {\em ICASSP}, 2023, pp. 1--5.

\bibitem{jiang2022dual}
Junkun Jiang, Jie Chen, and Yike Guo,
\newblock ``A dual-masked auto-encoder for robust motion capture with
  spatial-temporal skeletal token completion,''
\newblock in {\em Proceedings of the ACM International Conference on
  Multimedia}, 2022, pp. 5123--5131.

\bibitem{zhou2022quickpose}
Zhize Zhou, Qing Shuai, Yize Wang, Qi~Fang, Xiaopeng Ji, Fashuai Li, Hujun Bao,
  and Xiaowei Zhou,
\newblock ``Quickpose: Real-time multi-view multi-person pose estimation in
  crowded scenes,''
\newblock in {\em SIGGRAPH}, 2022, pp. 1--9.

\bibitem{pavlakos2019expressive}
Georgios Pavlakos, Vasileios Choutas, Nima Ghorbani, Timo Bolkart, Ahmed~AA
  Osman, Dimitrios Tzionas, and Michael~J Black,
\newblock ``Expressive body capture: {3D} hands, face, and body from a single
  image,''
\newblock in {\em CVPR}, 2019, pp. 10975--10985.

\bibitem{gong2022posetriplet}
Kehong Gong, Bingbing Li, Jianfeng Zhang, Tao Wang, Jing Huang, Michael~Bi Mi,
  Jiashi Feng, and Xinchao Wang,
\newblock ``Posetriplet: Co-evolving 3d human pose estimation, imitation, and
  hallucination under self-supervision,''
\newblock in {\em CVPR}, 2022, pp. 11017--11027.

\bibitem{xu2021graph}
Tianhan Xu and Wataru Takano,
\newblock ``Graph stacked hourglass networks for 3d human pose estimation,''
\newblock in {\em CVPR}, 2021, pp. 16105--16114.

\bibitem{zhao2022graformer}
Weixi Zhao, Weiqiang Wang, and Yunjie Tian,
\newblock ``Graformer: Graph-oriented transformer for 3d pose estimation,''
\newblock in {\em CVPR}, 2022, pp. 20438--20447.

\bibitem{pavllo20193d}
Dario Pavllo, Christoph Feichtenhofer, David Grangier, and Michael Auli,
\newblock ``3d human pose estimation in video with temporal convolutions and
  semi-supervised training,''
\newblock in {\em CVPR}, 2019, pp. 7753--7762.

\bibitem{zhang2022mixste}
Jinlu Zhang, Zhigang Tu, Jianyu Yang, Yujin Chen, and Junsong Yuan,
\newblock ``Mixste: Seq2seq mixed spatio-temporal encoder for 3d human pose
  estimation in video,''
\newblock in {\em CVPR}, 2022, pp. 13232--13242.

\bibitem{du2023flowpose}
Yaoyao Du, Zixiao Zhang, Zhihao Li, Peng Wei, Qingmin Liao, and Wenming Yang,
\newblock ``Flowpose: Conditional normalizing flows for 3d human pose and shape
  estimation from monocular videos,''
\newblock in {\em ICASSP}, 2023, pp. 1--5.

\bibitem{song2021denoising}
Jiaming Song, Chenlin Meng, and Stefano Ermon,
\newblock ``Denoising diffusion implicit models,''
\newblock in {\em ICLR}, 2021, pp. 1--1.

\bibitem{ho2020denoising}
Jonathan Ho, Ajay Jain, and Pieter Abbeel,
\newblock ``Denoising diffusion probabilistic models,''
\newblock {\em Advances in neural information processing systems}, pp.
  6840--6851, 2020.

\bibitem{gong2023diffpose}
Jia Gong, Lin~Geng Foo, Zhipeng Fan, Qiuhong Ke, Hossein Rahmani, and Jun Liu,
\newblock ``Diffpose: Toward more reliable 3d pose estimation,''
\newblock in {\em CVPR}, 2023, pp. 13041--13051.

\bibitem{choi2022diffupose}
Jeongjun Choi, Dongseok Shim, and H~Jin Kim,
\newblock ``Diffupose: Monocular 3d human pose estimation via denoising
  diffusion probabilistic model,''
\newblock {\em arXiv preprint arXiv:2212.02796}, 2022.

\bibitem{ionescu2013human3}
Catalin Ionescu, Dragos Papava, Vlad Olaru, and Cristian Sminchisescu,
\newblock ``Human3. 6m: Large scale datasets and predictive methods for 3d
  human sensing in natural environments,''
\newblock {\em IEEE Transactions on Pattern Analysis and Machine Intelligence},
  pp. 1325--1339, 2013.

\bibitem{mehta2017monocular}
Dushyant Mehta, Helge Rhodin, Dan Casas, Pascal Fua, Oleksandr Sotnychenko,
  Weipeng Xu, and Christian Theobalt,
\newblock ``Monocular 3d human pose estimation in the wild using improved cnn
  supervision,''
\newblock in {\em 3DV}, 2017, pp. 506--516.

\bibitem{li2020cascaded}
Shichao Li, Lei Ke, Kevin Pratama, Yu-Wing Tai, Chi-Keung Tang, and Kwang-Ting
  Cheng,
\newblock ``Cascaded deep monocular 3d human pose estimation with evolutionary
  training data,''
\newblock in {\em CVPR}, 2020, pp. 6173--6183.

\bibitem{li2022mhformer}
Wenhao Li, Hong Liu, Hao Tang, Pichao Wang, and Luc Van~Gool,
\newblock ``Mhformer: Multi-hypothesis transformer for 3d human pose
  estimation,''
\newblock in {\em CVPR}, 2022, pp. 13147--13156.

\bibitem{chen2018cascaded}
Yilun Chen, Zhicheng Wang, Yuxiang Peng, Zhiqiang Zhang, Gang Yu, and Jian Sun,
\newblock ``Cascaded pyramid network for multi-person pose estimation,''
\newblock in {\em CVPR}, 2018, pp. 7103--7112.

\end{thebibliography}

\end{document}